\documentclass[11pt]{article}

\usepackage[final]{acl}

\usepackage{times}
\usepackage{latexsym}
\usepackage{amsmath}

\usepackage{booktabs}
\usepackage{longtable}
\usepackage{supertabular}
\usepackage{threeparttable}
\usepackage{siunitx}  
\usepackage{multirow}

\usepackage{tabularx}   
\usepackage{caption}

\usepackage{fontspec}
\usepackage{polyglossia}

\setdefaultlanguage{english}
\setotherlanguage{russian}

\usepackage{microtype}

\usepackage{graphicx}

\usepackage{enumitem}

\title{mmPISA-bench: Do LLMs Reason Equally Well Across 43 Languages?}

\author{Yerzhan Sapenov \\
  Independent Scholar \\
  \texttt{ysapenov@gmail.com} \\\And
  Jaromir Savelka \\
  School of Computer Science, \\
  Carnegie Mellon University, \\
  Pittsburgh, PA, USA \\
  \texttt{jsavelka@cs.cmu.edu} \\}

\begin{document}
\maketitle

\begin{abstract}
We introduce \textit{mmPISA-bench}, a compact high-quality multilingual reasoning benchmark derived from the OECD Programme for International Student Assessment (PISA). The benchmark consists of 25 multiple-choice questions that require reasoning in order to be answered correctly. Each question is provided in official human translations to 43 languages and complemented with machine-translated counterparts (i.e., 2{,}150 data points in total). We evaluate two mainstream proprietary LLMs across languages, reasoning effort levels, and translation types in terms of their ability to answer the questions correctly. 
Our results show that modern LLMs can reason effectively across all evaluated languages, achieve accuracy comparable to human test-takers, with some performance variations across covered languages. We further find that machine-translated questions do not degrade accuracy relative to official human translations which suggests that high-quality machine translation~(synthetic data) might often be adequate for large-scale multilingual reasoning evaluations where official translations are not available. 
Finally, we analyze token usage and related inference cost and find that LLMs usage in some languages is simultaneously more expensive and less accurate.
\end{abstract}

\section{Introduction}

Large language models (LLMs) have demonstrated strong reasoning capabilities, yet their reasoning ability in many languages remains comparatively under-explored. 
Despite substantial investments in multilingual modeling, today’s LLM ecosystem is still largely dominated by English~\cite{wu-2025-bitter}. 
Persistent challenges include limited data for many languages, uneven performance across language communities, and tokenizer-induced disparities that can affect both effectiveness and cost \cite{qin-2025-survey}. 
As a result, reasoning in diverse languages remains at an early stage of evaluation and understanding \cite{ghosh-2025-survey}.

To study reasoning in many languages in a controlled setting, we leverage the OECD Programme for International Student Assessment~(PISA), which provides a rigorous framework for assessing student competencies and collects contextual data to explain performance differences \cite{oecd-2023-pisa22-framework}. 
PISA is a worldwide study that measures the proficiency of 15-year-old students in reading, mathematics, and science.
Crucially, PISA items undergo extensive translation and validation procedures to ensure cross-country comparability. 
This includes localization workflows (adaptation, translation, and validation), explicit translatability assessment, and reconciliation practices designed to reduce language-specific artifacts and improve equivalence across versions \cite{oecd-2024-pisa-techreport}. 
These properties make PISA questions a high-quality source for evaluating whether LLM performance on questions requiring reasoning is stable across languages, rather than being confounded by low-quality or inconsistent translations.

We investigate the following research questions:
\begin{enumerate}[label={(RQ\arabic*)},leftmargin=1.15cm, itemsep=0pt, topsep=1pt]
    \item How stable are frontier LLMs in answering questions requiring reasoning across many different languages?
    \item How does model performance on machine-translated questions compare to performance on official human translations?
    \item Do reasoning length vary systematically across languages, and how does this relate to accuracy?
\end{enumerate}

\noindent We release a dataset of 2,150 questions requiring reasoning drawn from PISA (25 multiple-choice questions in 43 languages with official human and matched machine translations). Further, we provide an analysis of stability and reasoning effort across human and machine translations in the 43 covered languages for selected frontier LLMs.

\section{Related Work}

\paragraph{PISA-based evaluation of LLMs.}
PISA questions have only been used sparingly for LLM evaluation, primarily because most items are not publicly available and access is restricted to a subset of released questions. \citet{takami-2023-PISA-japanese} evaluated ChatGPT using PISA multiple-choice questions, but limited the comparison to English and Japanese. 
\citet{basaran-2025-PISA-reading} focused exclusively on English reading items to assess reading proficiency. 
The most extensive prior use of PISA is PISA-Bench \cite{haller-2025-pisa}, which adapts PISA questions for evaluating vision language models; however, their benchmark relies on English source items that are machine-translated into five additional languages. In contrast, our work focuses on text-only reasoning and leverages official human translations across 43 languages.

\paragraph{Massively multilingual benchmarks.}
A broad range of benchmarks has been proposed to evaluate multilingual capabilities of LLMs. 
Global-MMLU \cite{singh-2025-globalMMLU} extends MMLU to 42 languages with a culturally sensitive subset, though a portion of the data is machine-translated.
M3Exam \cite{zhang-2023-m3exam} evaluates models across multiple modalities and difficulty levels in nine languages. 
MMLU-ProX \cite{xuan-2025-mmluprox} emphasizes reasoning complexity using translated items in 29 languages, while BUFFET \cite{asai-2024-buffet} unifies 15 tasks across 54 languages using machine-translated instructions. 
GlotEval \cite{luo-2025-gloteval} provides a framework for integrating and comparing results from 27 specialized multilingual benchmarks.

Several benchmarks target specific skills or modalities. Belebele \cite{bandarkar-2024-belebele} focuses on reading comprehension in 122 language variants based on FLORES-200 \cite{costajussa-2024-nllb}. 
\citet{shi-2023-reasoners} introduce a multilingual grade-school mathematics benchmark in 10 languages, demonstrating chain-of-thought reasoning beyond English. 
mSTEB \cite{beyene-2025-msteb} combines text and speech evaluation across many languages using FLORES-200 and FLEURS \cite{conneau-2022-fleurs}. 
In addition, AI Language Proficiency Monitor \cite{pomerenke-2025-monitor} aggregates results across multiple multilingual benchmarks to track model progress over time. 
Prior work has also examined trade-offs between accuracy and the language used for reasoning \cite{qi-2025-accuracy-cost}, as well as approaches to improve non-English reasoning efficiency \cite{huang-2024-mindmerger} or to disentangle language processing from reasoning \cite{zhao-2025-less}. 
Finally, identifying and covering low-resource languages remains an active area of research, with recent progress reaching over a thousand languages \cite{kargaran-2023-glotlid}.

Compared to these benchmarks, our dataset emphasizes high-quality, fully human-translated questions with explicit difficulty levels, enabling direct comparison to the performance of 15-year-old students across 43 languages.

\paragraph{Reasoning length and cost across languages.}
Recent studies indicate that multilingual differences in tokenization create systematic disparities in token counts across languages \cite{petrov-2023-tokenizers}.
For models that externalize their reasoning, these disparities manifest not only as higher cost but also as differences in \emph{reasoning length}. It is the amount of generated intermediate text used to justify an answer.
Such variation matters because longer reasoning traces increase inference cost and may reflect different internal strategies or difficulty in a given language.
While prior work has emphasized the cost implications of token inflation \cite{ahia-2023-cost}, we center our analysis on reasoning length itself and show that some languages elicit longer, more expensive reasoning while still exhibiting reduced accuracy.

\section{Dataset}

The collected dataset\footnote{https://github.com/ysapenov/mmPISA-bench} comprises 25 multiple-choice reasoning questions represented in 43 languages, derived from publicly available materials from the OECD Programme for International Student Assessment (PISA). 
Specifically, the collection includes 11 mathematics items from PISA 2022 and 14 reading comprehension items from PISA 2018 \cite{oecd-2026-pisaexamples}.

All publicly accessible PISA items were manually reviewed across available assessment years by the authors. 
The questions were hand-selected according to the following criteria:
\begin{enumerate}
    \item availability of a broad set of official language translations;
    \item exclusive reliance on textual information, excluding items requiring images or interactive components;
    \item multiple-choice format, excluding items that require evaluation of free-form responses.
\end{enumerate}

These constraints resulted in the inclusion of reading items from PISA 2018 and mathematics items from PISA 2022. Only languages for which complete translations existed for both reading and mathematics questions were retained. 
Question texts were obtained by structured scraping of interactive, language-specific subpages in \citealp{oecd-2026-pisaexamples}, followed by manual verification against the original English sources to ensure textual fidelity.

PISA items are annotated with eight difficulty levels, comprising six major levels, with level~1 further subdivided into three sublevels (1a, 1b, and 1c). 
The two assessed competencies are defined by PISA as follows. 
Mathematics is defined as students’ capacity to reason mathematically and to formulate, employ, and interpret mathematics to solve problems in a variety of real-world contexts, encompassing concepts, procedures, facts, and tools to describe, explain, and predict phenomena. 
Reading is defined as students’ capacity to understand, use, evaluate, reflect on, and engage with written texts in order to achieve goals, develop knowledge and potential, and participate in society \cite{oecd-2023-pisa22vol1}.

This massively multilingual dataset enables systematic evaluation of LLM performance on questions requiring reasoning across 43 languages. 
It also supports analysis of machine translation effects, as models can be evaluated on both official human translations and machine-translated variants of the same questions. 
For mathematics items, the availability of rationales further enables the construction of auxiliary or derived reasoning tasks. 
As shown in section~\ref{sec:5results}, model performance on machine-translated questions is not lower than on human-translated versions, suggesting that large-scale machine translation could be used to extend the dataset to hundreds of additional languages and to probe the breadth of multilingual reasoning capabilities.

Because each item is decomposed into context, question, and answer options with line-level consistency across languages, the dataset also enables controlled experiments involving mixed-language inputs at the component or line level. 
In the present experiments, models were not explicitly informed of the input language, leaving language identification implicit; providing such information may represent a potential avenue for improving performance.
Finally, the dataset supports extensions that increase task difficulty, such as introducing adversarial or incorrect answer options to study robustness across languages \cite{goral-2025-wrong-option}.

\section{Experimental Design}

Across all experiments, we issued a total of 107{,}500 API calls to the evaluated LLMs. Unless explicitly stated otherwise, temperature and all other model parameters were kept at their defaults.

The total number of evaluations is given by:
\begin{equation}
\label{eq:results}
\begin{split}
  & 107{,}500 \text{ data points} = \\
  &\; 25 \text{ questions} \times 43 \text{ languages} \\
  &\times 2 \text{ models} \times 2 \text{ translation types} \\
  &\times 5 \text{ reasoning effort levels} \times 5 \text{ repetitions}.
\end{split}
\end{equation}

Among the 25 questions, 11 assess mathematical reasoning and 14 assess reading comprehension. With respect to difficulty, 10 questions are at levels~1--2, 6 at levels~3--4, and 9 at levels~5--6, following the official PISA difficulty annotations \cite{oecd-2023-pisa22vol1}.

We evaluated two proprietary frontier LLMs, OpenAI's GPT and Anthropic Claude, under multiple reasoning effort settings. For GPT, we used the \texttt{GPT-5.1-2025-11-13} model and evaluated five effort configurations: \emph{none}, \emph{none with double prompt}, \emph{low}, \emph{medium}, and \emph{high}. The double-prompt, no-reasoning configuration was included to test the effect of prompt repetition, following the methodology of \citet{leviathan-2025-prompt}. The most recent \texttt{Opus-4-5-20251101} model does not support disabling reasoning effort. \texttt{Haiku-4-5-20251001} was used to approximate the no-reasoning setting, while higher effort levels were evaluated using the Opus model.

Two translation conditions were considered. The first uses the human translations provided by PISA. The second uses machine-translated versions produced with Google Translate. Because English and French both serve as source languages in the original PISA materials, machine-translated English and French items were obtained by translating each language into the other.

\begin{table}
\centering
\begin{threeparttable}
\normalsize
\begin{tabular}{lrrrr}
\toprule
& \multicolumn{2}{c}{\textbf{Claude}} & \multicolumn{2}{c}{\textbf{GPT}} \\
\cmidrule(lr){2-3} \cmidrule(lr){4-5}
\textbf{Language} & \textbf{none\tnote{a}} & \textbf{high} & \textbf{none} & \textbf{high} \\
\midrule
Albanian    & 88.0 & 94.4 & 80.0 & 98.4 \\
Arabic      & 86.4 & 98.4 & 81.6 & 94.4 \\
Azerbaijani & 84.0 & 95.2 & 76.0 & 96.0 \\
Basque      & 88.8 & 96.8 & 80.8 & 96.0 \\
Bokmål      & 85.6 & 96.0 & 76.0 & 99.2 \\
Bosnian     & 87.2 & 99.2 & 75.2 & 96.8 \\
Bulgarian   & 85.6 & 96.8 & 84.0 & 90.4 \\
Catalan     & 86.4 & 99.2 & 79.2 & 96.0 \\
Chinese     & 94.4 & 96.8 & 73.6 & 96.8 \\
Croatian    & 88.8 & 96.8 & 78.4 & 96.0 \\
Czech       & 84.8 & 96.8 & 79.2 & 94.4 \\
Danish      & 89.6 & 98.4 & 81.6 & 93.6 \\
Dutch       & 90.4 & 96.0 & 84.8 & 95.2 \\
English     & 90.4 & 100.0 & 76.8 & 96.8 \\
Estonian    & 89.6 & 96.0 & 75.2 & 93.6 \\
Finnish     & 88.0 & 100.0 & 83.2 & 98.4 \\
French      & 91.2 & 99.2 & 79.2 & 96.8 \\
Galician    & 88.8 & 98.4 & 74.4 & 93.6 \\
Georgian    & 84.8 & 100.0 & 80.0 & 99.2 \\
German      & 85.6 & 96.8 & 81.6 & 100.0 \\
Greek       & 84.0 & 88.0 & 79.2 & 90.4 \\
Hebrew      & 88.0 & 100.0 & 73.6 & 99.2 \\
Hungarian   & 80.8 & 98.4 & 84.8 & 98.4 \\
Icelandic   & 88.8 & 92.0 & 73.6 & 92.0 \\
Indonesian  & 85.6 & 95.2 & 76.0 & 96.0 \\
Italian     & 88.8 & 92.0 & 74.4 & 92.8 \\
Japanese    & 88.0 & 99.2 & 78.4 & 96.8 \\
Kazakh      & 80.0 & 93.6 & 74.4 & 92.8 \\
Korean      & 88.0 & 92.0 & 73.6 & 93.6 \\
Latvian     & 86.4 & 96.0 & 77.6 & 96.8 \\
Lithuanian  & 91.2 & 95.2 & 80.8 & 97.6 \\
Malay       & 90.4 & 96.0 & 80.0 & 96.0 \\
Nynorsk     & 85.6 & 92.0 & 80.8 & 98.4 \\
Polish      & 87.2 & 94.4 & 78.4 & 96.0 \\
Portuguese  & 89.6 & 100.0 & 82.4 & 96.8 \\
Russian     & 85.6 & 95.2 & 84.8 & 94.4 \\
Serbian     & 87.2 & 99.2 & 77.6 & 96.0 \\
Slovak      & 88.0 & 96.0 & 74.4 & 95.2 \\
Slovenian   & 89.6 & 95.2 & 80.0 & 96.0 \\
Spanish     & 87.2 & 98.4 & 82.4 & 90.4 \\
Swedish     & 86.4 & 100.0 & 81.6 & 94.4 \\
Thai        & 85.6 & 94.4 & 84.0 & 96.8 \\
Turkish     & 92.0 & 100.0 & 80.8 & 96.8 \\
\midrule
\textbf{Total} & \textbf{87.5} & \textbf{96.6} & \textbf{78.9} & \textbf{95.7} \\
\bottomrule
\end{tabular}
\caption{Comparative Language Performance Accuracy, \%: Claude vs GPT.}
\label{tab:language-comparison}
\begin{tablenotes}
    \item[a] Haiku model was used for none reasoning case
\end{tablenotes}
\end{threeparttable}
\end{table}

Each question--language--model--configuration combination was evaluated independently five times to assess answer stability and estimate accuracy under stochastic generation. The system prompts used for both models are provided in Appendix~\ref{sec:prompts}. These prompts were restricted to enforcing a uniform multiple-choice answer format. Aside from the system prompt, no additional instructions or contextual information were supplied to the models. All evaluations were conducted in a zero-shot setting. Cost is computed as the sum of input tokens multiplied by the input token price and output tokens multiplied by the output token price.

\section{Results}
\label{sec:5results}
Subsections \ref{sec:rq1} and \ref{sec:rq3} report results on the original, human-translated questions only. 
Subsection \ref{sec:rq2} compares performance on original questions against their machine-translated counterparts.

\subsection{RQ1: Consistency across languages}
\label{sec:rq1}

Table~\ref{tab:language-comparison} reports accuracy by language for both models under no-reasoning and high-reasoning settings. 
Both models' performance varies somewhat across languages, and these differences persist across reasoning levels.
While performance differences across languages clearly exist the relatively bounded extent of the variations indicates reasonably capable multilingual reasoning behavior across the 43 studied languages.

\subsection{RQ2: Human vs machine translation}
\label{sec:rq2}

Table~\ref{tab:reasoning-effort} compares accuracy and token usage between original human translations and machine-translated questions across reasoning effort levels. 
Overall, accuracy on machine-translated questions is not lower than on the original versions for either model, and in several configurations it is marginally higher. 
This suggests that machine translation does not introduce  performance degradation for the evaluated reasoning tasks.

\begin{table}
\centering
\scriptsize 
\begin{tabular}{@{}ll rrr rrr@{}}
\toprule
& & \multicolumn{3}{c}{\textbf{Original}} & \multicolumn{3}{c}{\textbf{Machine}} \\
\cmidrule(l){3-5} \cmidrule(l){6-8}
\textbf{Model} & \textbf{Effort} & \textbf{Acc} & \textbf{In} & \textbf{Out} & \textbf{Acc} & \textbf{In} & \textbf{Out} \\
\midrule
\multirow{6}{*}{\textbf{Claude}} & high & 96.6 & 4219 & 2258 & 97.4 & 4101 & 2230 \\
                                 & medium & 96.2 & 4767 & 796 & 96.6 & 4650 & 778 \\
                                 & low & 94.6 & 4767 & 515 & 95.5 & 4650 & 513 \\
                                 & double & 86.6 & 8304 & 2708 & 87.0 & 8068 & 2708 \\
                                 & none & 87.5 & 4219 & 2822 & 87.5 & 4101 & 2805 \\
\cmidrule(l){2-8}
                                 & \textbf{Total} & \textbf{92.3} & \textbf{26277} & \textbf{9098} & \textbf{92.8} & \textbf{25569} & \textbf{9034} \\
\midrule
\multirow{6}{*}{\textbf{GPT}} & high & 95.7 & 3087 & 3709 & 96.1 & 2974 & 3368 \\
                              & medium & 94.7 & 3087 & 1434 & 94.9 & 2970 & 1325 \\
                              & low & 93.7 & 3087 & 582 & 94.3 & 2970 & 562 \\
                              & double & 84.2 & 6089 & 59 & 85.5 & 5854 & 59 \\
                              & none & 78.9 & 3087 & 59 & 79.8 & 2970 & 59 \\
\cmidrule(l){2-8}
                              & \textbf{Total} & \textbf{89.5} & \textbf{18439} & \textbf{5843} & \textbf{90.1} & \textbf{17738} & \textbf{5374} \\
\bottomrule
\end{tabular}
\caption{Model accuracy in \%, thousands of input and output tokens across model reasoning effort levels.}
\label{tab:reasoning-effort}
\end{table}

Table~\ref{tab:difficulty} further breaks down results by PISA difficulty level. 
As expected, accuracy decreases with increasing difficulty, while average token usage per question increases. 
This trend is consistent across both original and machine-translated questions, indicating that difficulty effects dominate translation effects.

\begin{table}
\centering
\scriptsize 
\begin{tabular}{@{}ll rrr rrr@{}}
\toprule
& & \multicolumn{3}{c}{\textbf{Original}} & \multicolumn{3}{c}{\textbf{Machine}} \\
\cmidrule(l){3-5} \cmidrule(l){6-8}
\textbf{Model} & \textbf{Difficulty} & \textbf{Acc} & \textbf{In} & \textbf{Out} & \textbf{Acc} & \textbf{In} & \textbf{Out} \\
\midrule
\multirow{3}{*}{\textbf{Claude}} & level 5-6 & 80.0 & 869 & 415 & 80.5 & 843 & 411 \\
                                 & level 3-4 & 99.1 & 952 & 302 & 99.6 & 931 & 305 \\
                                 & level 1-2 & 99.3 & 1092 & 292 & 99.7 & 1061 & 288 \\
\midrule                                 
\multirow{3}{*}{\textbf{GPT}} & level 5-6 & 75.6 & 610 & 475 & 76.7 & 586 & 429 \\
                              & level 3-4 & 93.9 & 683 & 112 & 94.8 & 659 & 117 \\
                              & level 1-2 & 99.2 & 756 & 49  & 99.4 & 727 & 44  \\
\bottomrule
\end{tabular}
\caption{Model accuracy in \%, average token usage per question across difficulty levels.}
\label{tab:difficulty}
\end{table}

Table~\ref{tab:category-types} shows results by question category. 
Both models perform better on reading comprehension than on mathematics, with similar patterns observed for original and machine-translated inputs.
Token usage differs substantially between categories, particularly the GPT model.

\begin{table}
\centering
\scriptsize 
\begin{tabular}{@{}ll rrr rrr@{}}
\toprule
& & \multicolumn{3}{c}{\textbf{Original}} & \multicolumn{3}{c}{\textbf{Machine}} \\
\cmidrule(l){3-5} \cmidrule(l){6-8}
\textbf{Model} & \textbf{Category} & \textbf{Acc} & \textbf{In} & \textbf{Out} & \textbf{Acc} & \textbf{In} & \textbf{Out} \\
\midrule
\multirow{2}{*}{\textbf{Claude}} & Math & 85.1 & 557 & 355 & 84.9 & 542 & 357 \\
                                 & Reading & 98.0 & 1308 & 325 & 98.9 & 1273 & 320 \\
\midrule
\multirow{2}{*}{\textbf{GPT}} & Math & 82.4 & 400 & 372 & 83.4 & 381 & 344 \\
                              & Reading & 95.0 & 911 & 96 & 95.4 & 879 & 86 \\
\bottomrule
\end{tabular}
\caption{Model accuracy in \%, average token usage per question across categories.}
\label{tab:category-types}
\end{table}

\subsection{RQ3: Reasoning length}
\label{sec:rq3}

Table~\ref{tab:total} summarizes the overall accuracy--cost trade-off.
Claude achieves higher average accuracy, but at more than double the cost of GPT.

\begin{table}
\centering
\begin{tabular}{lrr}
\toprule
\textbf{Model} & \textbf{Accuracy} & \textbf{Cost (\$)} \\ 
\midrule
Claude         & 92.3\%            & 198.1         \\ 
GPT            & 89.5\%            &  81.5         \\ 
\bottomrule
\end{tabular}
\caption{Comparison of Model Accuracy and Cost.}
\label{tab:total}
\end{table}

\begin{figure*}[!htbp]
  \includegraphics[width=0.48\linewidth]{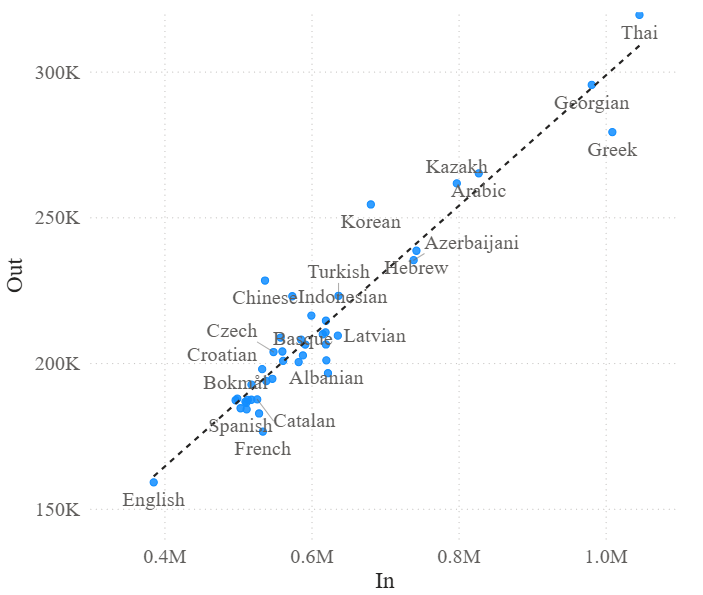} \hfill
  \includegraphics[width=0.48\linewidth]{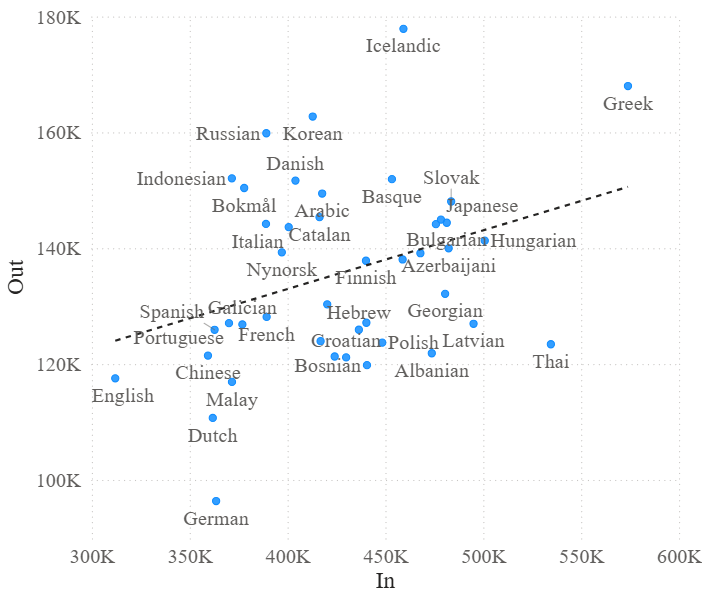} 
  \caption {Claude (left) and GPT (right) input and output tokens.}
  \label{fig:in-out}
\end{figure*}

Figure~\ref{fig:in-out} illustrates the relationship between input and output token usage across languages. 
Claude exhibits a strong positive correlation between input and output tokens, whereas this relationship is notably weaker for GPT.
The correlation between input and output tokens is extremely high for Claude (0.950) but much weaker for GPT~(0.334).
This indicates that Claude’s reasoning length scales closely with the length of the input across languages, while GPT exhibits more decoupled behavior in which longer prompts do not consistently result in longer generated reasoning.

\begin{figure*}[!htbp]
  \includegraphics[width=0.48\linewidth]{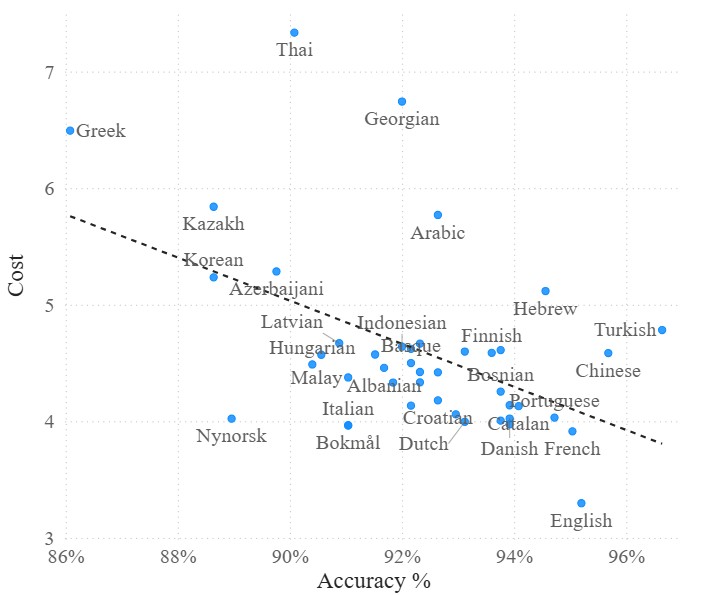} \hfill
  \includegraphics[width=0.48\linewidth]{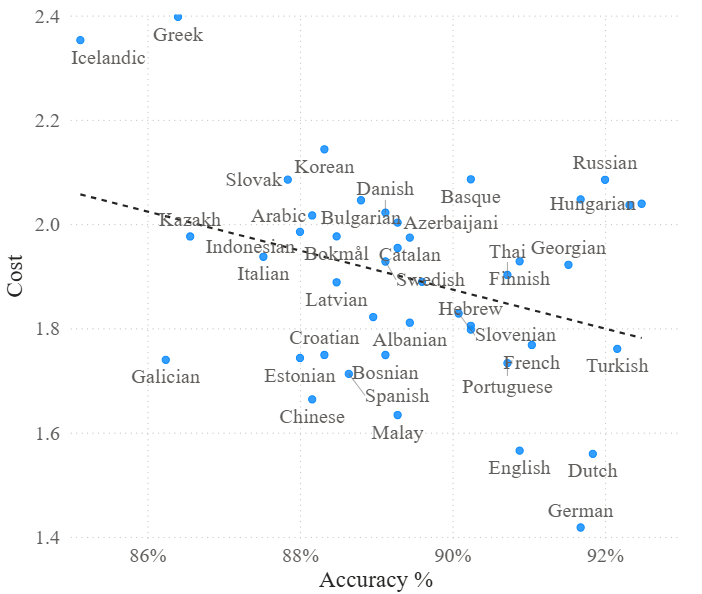} 
  \caption {Claude (left) and GPT (right) accuracy (\%) versus cost (\$).}
  \label{fig:acc-cost}
\end{figure*}

Figure~\ref{fig:acc-cost} plots accuracy against cost by language. 
For both models, higher cost is generally associated with lower accuracy, yielding negative correlations (Claude: −0.484; GPT: −0.339). 
Importantly, cost reflects two distinct sources: tokenization-driven input inflation and variation in reasoning length (output tokens).
For Claude, languages with higher input token counts also tend to elicit longer reasoning traces (Figure~\ref{fig:in-out}), amplifying cost and contributing to lower accuracy.
In contrast, GPT exhibits weaker coupling between input length and output length, suggesting that cost–accuracy degradation cannot be attributed to tokenization alone but also reflects language-dependent differences in generated reasoning.

\subsection{Additional results}

\begin{figure*}[!htbp]
  \includegraphics[width=0.48\linewidth]{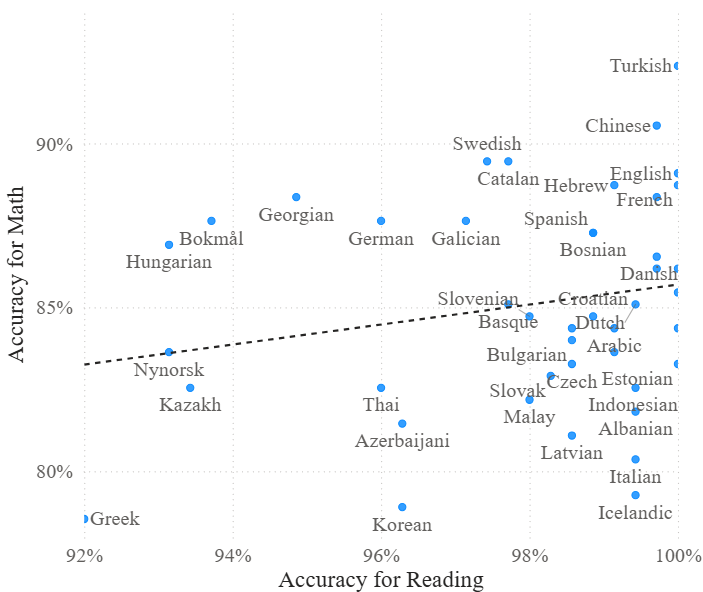} \hfill
  \includegraphics[width=0.48\linewidth]{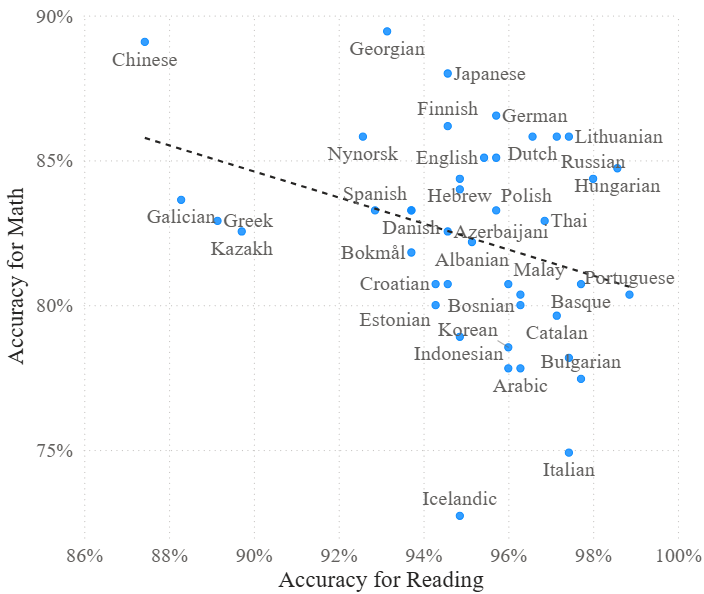} 
  \caption {Claude (left) and GPT (right) accuracy (\%) for Math and Reading categories.}
  \label{fig:math-read}
\end{figure*}

Figure~\ref{fig:math-read} compares accuracy between mathematics and reading questions. 
The correlation between category-specific accuracies for Claude is weak~(0.210), and for GPT it is negative (-0.324), suggesting that performance on one category does not reliably predict performance on the other.

Although LLM outputs are stochastic, qualitative inspection reveals that models can exhibit distinct reasoning behaviors across languages for the same question. 
Tables~\ref{tab:claude-comparison-23}, \ref{tab:claude-comparison-14}, and \ref{tab:claude-comparison-11} show that the model’s reasoning is not necessarily carried out in prompt language or English. 
In some cases, the input text is implicitly translated from the prompt language into another language used for reasoning.
Notably, for Kazakh prompts, Claude performs parts of its reasoning in Russian rather than English and may switch between languages within a single response. 
These behaviors were identifiable because one of the authors is fluent in Kazakh and Russian. 
Similar cross-language reasoning may occur in other languages but remain difficult to detect without native-language expertise.

\begin{table*}
\centering
\footnotesize
\begin{tabularx}{\textwidth}{X X}
\toprule
\textbf{Response A (Russian)} & \textbf{Response B (Kazakh)} \\ 
\midrule
Мне нужно определить, какие комментарии в этом форуме относятся к проблеме Индира\_88 (вопрос о том, можно ли давать аспирин курице). & 
Тауықтарға аспирин беру туралы сұрақ бойынша әрбір хабарламаның проблемасына қатысты немесе қатыстас емес екенін анықтау керек. \\
\addlinespace
1. НұргүлB79 - "Не знаю, безопасен ли аспирин для птиц...". Это прямое выступление по поводу проблемы Индиры. Иә (Да) & 
1. НұргүлB79 хабарламасы: "...сұрақты анықтамайды, бірақ қауіпсіздік мәселесін қозғайды" - Иә \\
\addlinespace
2. Мөлдір - "Я дал аспирин своей курице когда-то...". Это прямой опыт, связанный с проблемой Индиры. Иә (Да) & 
2. Мөлдір хабарламасы: "...БҰЛ ПРОБЛЕМАСЫНА ҚАТЫСТЫ (тәжірибе пен дозалау туралы)" - Иә \\
\addlinespace
3. Айдын\_ұсыныстар - "Не забудьте посмотреть мои отличные низкие цены...". Это спам/реклама, не относится к проблеме Индиры. Жоқ (Нет) & 
3. Айдын\_ұсыныстар хабарламасы: "ҚАТЫСТАС ЕМЕС (жөнсіз сату ойтарымы)" - Жоқ \\
\addlinespace
4. Бағылан - "Вы можете объяснить мне, как узнать, что курица больна?". Это новый вопрос, косвенно связанный... Жоқ (Нет) & 
4. Бағылан хабарламасы: "ПРОБЛЕМАСЫНА ҚАТЫСТАС ЕМЕС (өзгеше сұрақ)" - Жоқ \\
\addlinespace
5. Фархат - "Привет Индира, я ветеринар...". Это прямой профессиональный ответ на вопрос Индиры. Иә (Да) & 
5. Фархат хабарламасы: "БҰЛ ПРОБЛЕМАСЫНА ТІКЕЛЕЙ ҚАТЫСТЫ (түсінік дәрігерден)" - Иә \\
\addlinespace
Получается: ['Иә', 'Иә', 'Жоқ', 'Жоқ', 'Иә'] & 
Сонымен: ['Иә', 'Иә', 'Жоқ', 'Жоқ', 'Иә'] \\ 
\bottomrule
\end{tabularx}
\caption{Comparison of Claude reasoning for the same question \#23 in Kazakh language across independent runs (same prompt, identical settings).}
\label{tab:claude-comparison-23}
\end{table*}

\section{Discussion}

This section interprets the empirical results in light of the three research questions, focusing on multilingual reasoning robustness, translation effects, and cost--accuracy trade-offs.

\subsection{RQ1: Consistency across languages}

The results indicate that leading LLMs can reason across all evaluated languages at a level comparable to that expected of 15-year-old students.
Compared to earlier evaluations, cross-lingual performance gaps appear reduced, suggesting improved multilingual robustness in recent models \cite{petrov-2023-tokenizers}. 
As shown in Table~\ref{tab:language-comparison}, under high reasoning effort GPT achieves 100\% accuracy only for German, whereas Claude reaches perfect accuracy in seven languages. 

Claude’s accuracy spans 88\%--100\% under high effort and 80\%--94.4\% without explicit reasoning, corresponding to ranges of 12.0\%--14.4\%, respectively. 
GPT exhibits narrower ranges: 90.4\%--100\% under high effort and 73.6\%--84.8\% without reasoning, corresponding to ranges of 9.6\%--11.2\%. 
Thus, although Claude achieves higher average accuracy, GPT displays less variability across languages.

\subsection{RQ2: Human vs machine translation}

Prompt repetition improves accuracy for GPT in the no-reasoning setting, as shown in Table~\ref{tab:reasoning-effort}. 
In contrast, Claude Haiku does not benefit from double prompting, consistent with prior findings~\cite{leviathan-2025-prompt}. 
These results suggest that prompt repetition is model-dependent and primarily beneficial for architectures that do not expose internal reasoning by default.

Increasing question difficulty leads to lower accuracy and higher output token usage, as shown in Table~\ref{tab:difficulty}. 
Notably, higher-difficulty questions are often shorter in terms of input tokens, yet they elicit longer outputs, indicating more elaborate reasoning processes that more closely resemble human problem-solving behavior.

Table~\ref{tab:category-types} further shows category-specific differences.
GPT uses substantially more output tokens for mathematics questions than for reading, whereas Claude’s output token usage is comparatively similar across categories. 
This suggests different internal strategies for handling numerical reasoning versus textual comprehension.

\subsection{RQ3: Reasoning length}

Token usage varies substantially across languages, corroborating earlier observations disparities based on tokenization premium \cite{petrov-2023-tokenizers}. 
For Claude, the highest tokenization premium is observed for Thai in Figure~\ref{fig:in-out}, with factors of 2.71 for input tokens and 2.01 for output tokens relative to English. 
Crucially, we also observe systematic variation in reasoning length across languages, operationalized as output tokens under a fixed reasoning-effort setting. For some languages, models produce substantially longer rationales even when answering the same items.
These values are lower than previously reported maxima for the \texttt{cl100k\_base} tokenizer \cite{petrov-2023-tokenizers}, indicating partial mitigation of extreme token inflation.

For GPT, the largest input token premium occurs for Greek (1.84), while the largest output token premium is observed for Icelandic (1.51) in Figure~\ref{fig:in-out}. 
In the Claude case, English consistently yields the lowest token usage and exhibits a strong correlation between input and output tokens, suggesting that languages that are “longer to read” also tend to elicit longer reasoning traces. 
In contrast, GPT shows weaker coupling between input and output length, indicating that cross-linguistic differences in generated reasoning verbosity are not fully explained by tokenization alone.

Figure~\ref{fig:acc-cost} illustrates the relationship between cost and accuracy. 
For both models, languages that are less costly to process tend to yield higher accuracy. For example, Claude incurs 2.22$\times$ higher cost on Thai than on English while achieving 5.1 percentage points lower accuracy.
Similarly, GPT spends 1.69$\times$ more on Greek than on German, with a corresponding 5.3 percentage point accuracy decrease. 
These findings indicate that some languages are simultaneously more expensive and less accurate, reinforcing the importance of cost-aware multilingual evaluation.
Taken together, these results suggest that multilingual evaluation should report not only accuracy but also reasoning length, since some languages systematically induce longer and sometimes less effective reasoning.

\subsection{Additional results}

Figure~\ref{fig:math-read} compares performance on mathematics and reading questions. 
The weak correlation between category-specific accuracies suggests that strong performance in one category does not necessarily transfer to the other. 
Notably, GPT achieves one of its lowest reading accuracies overall, yet performs strongly on reading questions in Chinese, highlighting language--category interactions that merit further investigation.

The qualitative examples in Tables~\ref{tab:claude-comparison-23}, \ref{tab:claude-comparison-11} and \ref{tab:claude-comparison-14} show that multilingual differences extend beyond accuracy and reasoning length to the linguistic behavior of model reasoning. In particular, Claude exhibits variation in reasoning structure across languages, including explicit translation during reasoning and cross-language reasoning. Notably, in some cases the model switches to a language different from both the input language and English~(e.g., Russian when prompted in Kazakh), suggesting that intermediate reasoning may occur in a latent pivot language. These behaviors were identifiable only because one of the authors is fluent in Kazakh and Russian, highlighting a broader evaluation challenge: such phenomena may remain invisible without native-language expertise. This suggests that multilingual LLM evaluation would benefit from qualitative inspection in addition to aggregate accuracy metrics.

\section{Conclusion}

Our results show that leading proprietary LLMs are capable of reasoning across all 43 evaluated languages. 
While overall reasoning accuracy remains high, both performance and inference cost vary substantially across languages, reflecting differences in multilingual robustness and tokenization efficiency. 
These findings underscore the importance of evaluating LLM reasoning beyond English and of jointly considering accuracy and cost when assessing multilingual capabilities.

\section{Future Work}

Several directions follow naturally from this study. 
First, the dataset can be extended to a substantially larger number of languages by leveraging machine translation systems, such as Google Translate, which support more than 250 languages. 
Targeted human translations could be obtained for selected ultra low-resource languages to assess reasoning capabilities under extreme data scarcity. Second, future work may compare the observed trends with those of open-source LLMs, enabling analysis of how architectural choices and training regimes affect multilingual reasoning. Finally, a longitudinal analysis tracking multilingual reasoning performance of LLMs over the past several years would provide insight into the pace and nature of progress in this area.

\section*{Acknowledgments}

We thank the Organisation for Economic Co-operation and Development (OECD) for conducting the PISA assessments worldwide and for providing open access to the test questions in numerous languages. The second author acknowledges generous support of Carnegie Mellon-Accenture Center of Excellence in AI-Enabled Workforce Training (ACE-AI). The content of this paper does not necessarily reflect the position or the policy of the funding organization and no official endorsement should be inferred.

\bibliography{main}

@inproceedings{zhao-2025-less,
    title = {{When Less Language is More: Language-Reasoning Disentanglement Makes LLMs Better Multilingual Reasoners}},
    author = {Zhao, Weixiang and 
      Guo, Jiahe and 
      Deng, Yang and 
      Wu, Tongtong and 
      Zhang, Wenxuan and 
      Hu, Yulin and 
      Sui, Xingyu and 
      Zhao, Yanyan and 
      Che, Wanxiang and 
      Qin, Bing and 
      Chua, Tat-Seng and 
      Liu, Ting},
    booktitle = {Advances in Neural Information Processing Systems (NeurIPS)},
    year = {2025},
    url = {https://arxiv.org/abs/2505.15257}
}

@inproceedings{goral-2025-wrong-option,
    title = {{Wait, that's not an option: LLMs Robustness with Incorrect Multiple-Choice Options}},
    author = {G{\'o}ral, Gracjan and 
      Wi{\'s}nios, Emilia and 
      Sankowski, Piotr and 
      Budzianowski, Pawe{\l}},
    booktitle = {Proceedings of the 63rd Annual Meeting of the Association for Computational Linguistics (Volume 1: Long Papers)},
    year = {2025},
    url = {https://aclanthology.org/2025.acl-long.75},
    pages = {1495--1515}
}

@article{leviathan-2025-prompt,
    title = {{Prompt Repetition Improves Non-Reasoning LLMs}},
    author = {Leviathan, Yaniv and 
              Kalman, Matan and 
              Matias, Yossi},
    journal = {arXiv preprint arXiv:2512.14982},
    year = {2025},
    url = {https://arxiv.org/abs/2512.14982},
    doi = {10.48550/arXiv.2512.14982},
    month = {dec}
}

@inproceedings{ghosh-2025-survey,
    title = {{A Survey of Multilingual Reasoning in Language Models}},
    author = {Ghosh, Akash and 
      Datta, Debayan and 
      Saha, Sriparna and 
      Agarwal, Chirag},
    booktitle = {Proceedings of the 2025 Conference on Empirical Methods in Natural Language Processing (EMNLP)},
    year = {2025},
    url = {https://arxiv.org/abs/2502.09457}
}

@inproceedings{kargaran-2023-glotlid,
    title = {{GlotLID: Language Identification for Low-Resource Languages}},
    author = {Kargaran, Amir Hossein and 
      Imani, Ayyoob and 
      Yvon, Fran{\c{c}}ois and 
      Sch{\"u}tze, Hinrich},
    booktitle = {Proceedings of the 2023 Conference on Empirical Methods in Natural Language Processing (EMNLP)},
    year = {2023},
    url = {https://arxiv.org/abs/2310.16248}
}

@inproceedings{huang-2024-mindmerger,
    title = {{MindMerger: Efficiently Boosting LLM Reasoning in non-English Languages}},
    author = {Huang, Zixian and 
      Zhu, Wenhao and 
      Cheng, Gong and 
      Li, Lei and 
      Yuan, Fei},
    booktitle = {Advances in Neural Information Processing Systems (NeurIPS)},
    year = {2024},
    url = {https://proceedings.neurips.cc/paper_files/paper/2024/hash/3bf80b34f731313b8292f4578e820c90-Abstract-Conference.html}
}

@inproceedings{qi-2025-accuracy-cost,
    title = {{When Models Reason in Your Language: Controlling Thinking Language Comes at the Cost of Accuracy}},
    author = {Qi, Jirui and 
      Chen, Shan and 
      Xiong, Zidi and 
      Fern{\'a}ndez, Raquel and 
      Bitterman, Danielle S. and 
      Bisazza, Arianna},
    booktitle = {Findings of the Association for Computational Linguistics: EMNLP 2025},
    year = {2025},
    url = {https://aclanthology.org/2025.findings-emnlp.1103},
    pages = {20279--20296}
}

@article{pomerenke-2025-monitor,
    title = {{The AI Language Proficiency Monitor – Tracking the Progress of LLMs on Multilingual Benchmarks}},
    author = {Pomerenke, David and 
      Nothnagel, Jonas and 
      Ostermann, Simon},
    journal = {arXiv preprint arXiv:2507.08538},
    year = {2025},
    url = {https://arxiv.org/abs/2507.08538}
}

@article{costajussa-2024-nllb,
    title = {{Scaling neural machine translation to 200 languages}},
    author = {Costa-juss{\`a}, Marta Ruiz and 
      Cross, James and 
      {\c{C}}elebi, Onur and 
      Elbayad, Maha and 
      Heafield, Kenneth and 
      Heffernan, Kevin and 
      Kalbassi, Elahe and 
      Lam, Janice and 
      Licht, Daniel and 
      Maillard, Jean and 
      Sun, Anna and 
      Wang, Skyler and 
      Wenzek, Guillaume and 
      Youngblood, Al and 
      Akula, Bapi and 
      Barrault, Loic and 
      Mejia Gonzalez, Gabriel and 
      Hansanti, Prangthip and 
      Hoffman, John and 
      Jarrett, Semarley and 
      Sadagopan, Kaushik Ram and
      Rowe, Dirk and
      Spruit, Shannon and
      Tran, Chau and
      Andrews, Pierre and
      Ayan, Necip Fazil and
      Bhosale, Shruti and
      Edunov, Sergey and
      Fan, Angela and
      Gao, Cynthia and
      Goswami, Vedanuj and
      Guzmán, Francisco and
      Koehn, Philipp and
      Mourachko, Alexandre and
      Ropers, Christophe and
      Saleem, Safiyyah and
      Schwenk, Holger and
      Wang, Jeff},
    journal = {Nature},
    volume = {630},
    number = {8018},
    pages = {841--846},
    year = {2024},
    url = {https://www.nature.com/articles/s41586-024-07335-x}
}

@article{qin-2025-survey,
    title = {{A survey of multilingual large language models}},
    author = {Qin, Libo and 
      Chen, Qiguang and 
      Zhou, Yuhang and 
      Chen, Zhi and 
      Li, Yinghui and 
      Liao, Lizi and 
      Li, Min and 
      Che, Wanxiang and 
      Yu, Philip S.},
    journal = {Patterns},
    volume = {6},
    number = {1},
    pages = {101118},
    year = {2025},
    publisher = {Elsevier},
    url = {https://doi.org/10.1016/j.patter.2024.101118}
}

@inproceedings{beyene-2025-msteb,
    title = {{mSTEB: Massively Multilingual Evaluation of LLMs on Speech and Text Tasks}},
    author = {Beyene, Luel Hagos and 
      Verma, Vivek and 
      Ma, Min and 
      Alabi, Jesujoba O. and 
      Schmidt, Fabian David and 
      Nakatumba-Nabende, Joyce and 
      Adelani, David Ifeoluwa},
    booktitle = {Proceedings of the 2025 IEEE Automatic Speech Recognition and Understanding Workshop (ASRU)},
    year = {2025},
    url = {https://arxiv.org/abs/2506.08400}
}

@inproceedings{conneau-2022-fleurs,
  title={{FLEURS: Few-shot Learning Evaluation of Universal Representations of Speech}},
  author={Conneau, Alexis and Ma, Min and Khanuja, Simran and Zhang, Yu and Axelrod, Vera and Dalmia, Siddharth and Riesa, Jason and Rivera, Clara and Bapna, Ankur},
  booktitle={Proceedings of the 2022 IEEE Spoken Language Technology Workshop (SLT)},
  pages={798--805},
  year={2022},
  organization={IEEE},
  url={https://api.semanticscholar.org/CorpusID:249062909}
}

@article{wu-2025-bitter,
    title = {{The Bitter Lesson Learned from 2,000+ Multilingual Benchmarks}},
    author = {Wu, Minghao and 
      Wang, Weixuan and 
      Liu, Sinuo and 
      Yin, Huifeng and 
      Wang, Xintong and 
      Zhao, Yu and 
      Lyu, Chenyang and 
      Wang, Longyue and 
      Luo, Weihua and 
      Zhang, Kaifu},
    journal = {arXiv preprint arXiv:2504.15521},
    year = {2025},
    url = {https://arxiv.org/abs/2504.15521}
}

@inproceedings{shi-2023-reasoners,
    title = {{Language Models are Multilingual Chain-of-Thought Reasoners}},
    author = {Shi, Freda and 
      Suzgun, Mirac and 
      Freitag, Markus and 
      Wang, Xuezhi and 
      Srivats, Suraj and 
      Vosoughi, Soroush and 
      Chung, Hyung Won and 
      Tay, Yi and 
      Ruder, Sebastian and 
      Zhou, Denny and 
      Das, Dipanjan and 
      Wei, Jason},
    booktitle = {Proceedings of the 11th International Conference on Learning Representations (ICLR)},
    year = {2023},
    url = {https://openreview.net/forum?id=fR3wGCk-IXp}
}

@inproceedings{bandarkar-2024-belebele,
    title = {{The Belebele Benchmark: a Parallel Reading Comprehension Dataset in 122 Language Variants}},
    author = {Bandarkar, Lucas and 
      Liang, Davis and 
      Muller, Benjamin and 
      Artetxe, Mikel and 
      Shukla, Satya Narayan and 
      Husa, Donald and 
      Goyal, Naman and 
      Krishnan, Abhinandan and 
      Zettlemoyer, Luke and 
      Khabsa, Madian},
    booktitle = {Proceedings of the 62nd Annual Meeting of the Association for Computational Linguistics (Volume 1: Long Papers)},
    month = aug,
    year = {2024},
    address = {Bangkok, Thailand},
    publisher = {Association for Computational Linguistics},
    url = {https://aclanthology.org/2024.acl-long.44},
    pages = {749--775}
}

@inproceedings{luo-2025-gloteval,
    title = {{GlotEval: A Test Suite for Massively Multilingual Evaluation of Large Language Models}},
    author = {Luo, Hengyu and 
      Li, Zihao and 
      Attieh, Joseph and 
      Devkota, Sawal and 
      de Gibert, Ona and 
      Huang, Xu and 
      Ji, Shaoxiong and 
      Lin, Peiqin and 
      Mantina, Bhavani Sai Praneeth Varma and 
      Sreenidhi, Ananda and 
      V{\'{a}}zquez, Ra{\'{u}}l and 
      Wang, Mengjie and 
      Yusofi, Samea and 
      Yuan, Fei and 
      Tiedemann, J{\"{o}}rg},
    booktitle = {Proceedings of the 2025 Conference on Empirical Methods in Natural Language Processing: System Demonstrations},
    year = {2025},
    url = {https://aclanthology.org/2025.emnlp-demos.43},
    pages = {602--614}
}

@inproceedings{asai-2024-buffet,
    title = {{BUFFET: Benchmarking Large Language Models for Few-shot Cross-lingual Transfer}},
    author = {Asai, Akari and 
      Kudugunta, Sneha and 
      Yu, Xinyan Velocity and 
      Blevins, Terra and 
      Gonen, Hila and 
      Reid, Machel and 
      Tsvetkov, Yulia and 
      Ruder, Sebastian and 
      Hajishirzi, Hannaneh},
    booktitle = {Proceedings of the 2024 Conference of the North American Chapter of the Association for Computational Linguistics: Human Language Technologies (Volume 1: Long Papers)},
    year = {2024},
    url = {https://aclanthology.org/2024.naacl-long.100},
    pages = {1771--1800}
}

@article{xuan-2025-mmluprox,
    title = {{MMLU-ProX: A Multilingual Benchmark for Advanced Large Language Model Evaluation}},
    author = {Xuan, Weihao and 
      Yang, Rui and 
      Qi, Heli and 
      Zeng, Qingcheng and 
      Xiao, Yunze and 
      Feng, Aosong and 
      Liu, Dairui and 
      Xing, Yun and 
      Wang, Junjue and 
      Gao, Fan and 
      Lu, Jinghui and 
      Jiang, Yuang and 
      Li, Huitao and 
      Li, Xin and 
      Yu, Kunyu and 
      Dong, Ruihai and 
      Gu, Shangding and 
      Li, Yuekang and 
      Xie, Xiaofei and 
      Juefei-Xu, Felix and 
      Khomh, Foutse and 
      Yoshie, Osamu and 
      Chen, Qingyu and 
      Teodoro, Douglas and 
      Liu, Nan and 
      Goebel, Randy and 
      Ma, Lei and 
      Marrese-Taylor, Edison and 
      Lu, Shijian and 
      Iwasawa, Yusuke and 
      Matsuo, Yutaka and 
      Li, Irene},
    journal = {arXiv preprint arXiv:2503.10497},
    year = {2025},
    url = {https://arxiv.org/abs/2503.10497}
}

@inproceedings{zhang-2023-m3exam,
    title = {{M3Exam: A Multilingual, Multimodal, Multilevel Benchmark for Examining Large Language Models}},
    author = {Zhang, Wenxuan and 
      Aljunied, Sharifah Mahani and 
      Gao, Chang and 
      Chia, Yew Ken and 
      Bing, Lidong},
    booktitle = {Advances in Neural Information Processing Systems 36 (NeurIPS)},
    year = {2023},
    url = {https://arxiv.org/abs/2306.05179}
}

@inproceedings{singh-2025-globalMMLU,
    title = {{Global MMLU: Understanding and Addressing Cultural and Linguistic Biases in Multilingual Evaluation}},
    author = {Singh, Shivalika and 
              Romanou, Angelika and 
              Fourrier, Cl{\'e}mentine and 
              Adelani, David Ifeoluwa and 
              Ngui, Jian Gang and 
              Vila-Suero, Daniel and 
              Limkonchotiwat, Peerat and 
              Marchisio, Kelly and 
              Leong, Wei Qi and 
              Ng, Raymond and 
              Longpre, Shayne and 
              Oh, Alice and 
              Martins, Andre F. T. and 
              Choshen, Leshem and 
              Ippolito, Daphne and 
              Ferrante, Enzo and 
              Fadaee, Marzieh and
              Ermis, Beyza  and
              Hooker, Sara},
    booktitle = {Proceedings of the 63rd Annual Meeting of the Association for Computational Linguistics (Volume 1: Long Papers)},
    month = {jul},
    year = {2025},
    address = {Vienna, Austria},
    publisher = {Association for Computational Linguistics},
    url = {https://aclanthology.org/2025.acl-long.919},
    pages = {18761--18799}
}

@article{haller-2025-pisa,
    title = {{PISA-Bench: The PISA Index as a Multilingual and Multimodal Metric for the Evaluation of Vision-Language Models}},
    author = {Haller, Patrick and 
      Barth, Fabio and 
      Golde, Jonas and 
      Rehm, Georg and 
      Akbik, Alan},
    journal = {arXiv preprint arXiv:2510.24792},
    year = {2025},
    url = {https://arxiv.org/abs/2510.24792}
}

@inproceedings{takami-2023-PISA-japanese,
    title = {{Exploring ChatGPT Performance on PISA Multiple Choice Sample Questions Comparing English and Japanese Expression}},
    author = {Takami, Kyosuke},
    booktitle = {Proceedings of the Workshop on The Applications of Generative Artificial Intelligence (GAI) in Education, 31st International Conference on Computers in Education (ICCE)},
    year = {2023},
    address = {Matsue, Japan},
    url = {https://www.researchgate.net/publication/376173798_Exploring_ChatGPT_Performance_on_PISA_Multiple_Choice_Sample_Questions_Comparing_English_and_Japanese_Expression}
}

@article{basaran-2025-PISA-reading,
    title = {{Assessing AI in Educational Evaluation: A Comprehensive Analysis of ChatGPT's Performance on PISA Reading Skills}},
    author = {Ba{\c{s}}aran, Mehmet and 
              Vural, {\"O}mer Faruk and 
              Tand{\i}rc{\i}, Cennet},
    journal = {Technology, Knowledge and Learning},
    year = {2025},
    doi = {10.1007/s10758-025-09883-1},
    url = {https://doi.org/10.1007/s10758-025-09883-1},
    publisher = {Springer}
}

@inproceedings{petrov-2023-tokenizers,
    title = {{Language Model Tokenizers Introduce Unfairness Between Languages}},
    author = {Petrov, Aleksandr and 
              La Malfa, Emanuele and 
              Torr, Philip H. S. and 
              Bibi, Adel},
    booktitle = {{Proceedings of the 37th International Conference on Neural Information Processing Systems (NeurIPS 2023)}},
    year = {2023},
    url = {https://dl.acm.org/doi/10.5555/3666122.3667730},
    pages = {36963--36990},
    publisher = {Curran Associates Inc.}
}

@inproceedings{ahia-2023-cost,
    title = {{Do All Languages Cost the Same? Tokenization in the Era of Commercial Language Models}},
    author = {Ahia, Orevaoghene and 
              Kumar, Sachin and 
              Gonen, Hila and 
              Kasai, Jungo and 
              Mortensen, David R. and 
              Smith, Noah A. and 
              Tsvetkov, Yulia},
    booktitle = {{Proceedings of the 2023 Conference on Empirical Methods in Natural Language Processing}},
    month = {dec},
    year = {2023},
    address = {Singapore},
    publisher = {Association for Computational Linguistics},
    url = {https://aclanthology.org/2023.emnlp-main.614},
    doi = {10.18653/v1/2023.emnlp-main.614},
    pages = {9904--9923}
}

@book{oecd-2024-pisa-techreport,
    title = {{PISA 2022 Technical Report}},
    author = {{OECD}},
    year = {2024},
    series = {{PISA}},
    publisher = {{OECD Publishing}},
    address = {Paris},
    doi = {10.1787/01820d6d-en},
    url = {https://doi.org/10.1787/01820d6d-en}
}

@book{oecd-2023-pisa22vol1,
    title = {{PISA 2022 Results (Volume I): The State of Learning and Equity in Education}},
    author = {{OECD}},
    year = {2023},
    series = {{PISA}},
    publisher = {{OECD Publishing}},
    address = {Paris},
    doi = {10.1787/53f23881-en},
    url = {https://doi.org/10.1787/53f23881-en}
}

@book{oecd-2023-pisa22-framework,
    title = {{PISA 2022 Assessment and Analytical Framework}},
    author = {{OECD}},
    year = {2023},
    series = {{PISA}},
    publisher = {{OECD Publishing}},
    address = {Paris},
    doi = {10.1787/dfe0bf9c-en},
    url = {https://doi.org/10.1787/dfe0bf9c-en}
}

@misc{oecd-2026-pisaexamples,
    title = {{PISA Test Examples: Mathematics and Reading}},
    author = {{OECD}},
    year = {2026},
    howpublished = {\url{https://www.oecd.org/en/about/programmes/pisa/pisa-test.html}},
    note = {Accessed: December 5, 2025}
}

\appendix

\section{Limitations}
\label{sec:limitations}

The evaluation in this study is limited to two proprietary large language models. 
While these models are representative of current frontier systems, their training data, architectures, and inference mechanisms are not publicly documented. 
Consequently, the observed patterns in multilingual reasoning, reasoning length, and cost may not generalize to other proprietary models or to open-source models developed under different training regimes.

In addition, the benchmark draws exclusively on publicly released PISA questions from specific assessment years (PISA 2018 for reading and PISA 2022 for mathematics). 
Although these items were selected to maximize language coverage and comparability, they represent only a subset of PISA competencies and formats. 
In particular, the exclusion of constructed-response items, visual prompts, and interactive tasks limits the scope of reasoning behaviors that can be evaluated. 
Extending the benchmark to additional PISA cycles or complementary assessment frameworks would help capture a wider spectrum of multilingual reasoning skills.

\section{System Prompts}
\label{sec:prompts}

The system prompt for GPT models: "Reply format: <LETTER>".
Interesting that the same prompt did not work for Claude. 
It looks like GPT can easily have hidden reasoning tokens, while Claude displays all the tokens.
So, the system prompt for Claude models:  "Reply text inside the <reasoning> tags. Output only the letter answer outside the tags."

\section{Additional Results}
\label{sec:results}

\begin{table*}
\centering
\footnotesize 
\begin{tabularx}{\textwidth}{X  X}
\toprule
\textbf{Response A} & \textbf{Response B} \\ 
\midrule
Looking at the data for Suður-Kórea (South Korea): 64.42, 64.08, 63.69 & 
Skoðum gögnin fyrir Suður-Kóreu: 64,42 (2005), 64,08 (2010), 63,69 (2015). \\
\addlinespace
Comparing to Panama: 64.33, 63.21, 62.11 & 
Samanborið við Panama: 64,33 (2005), 63,21 (2010), 62,11 (2015). \\
\addlinespace
In 2005: Suður-Kórea (64.42) > Panama (64.33) & 
Árið 2005: Suður-Kórea (64,42) > Panama (64,33) - já, Suður-Kórea hæst \\
\addlinespace
In 2010: Suður-Kórea (64.08) > Panama (63.21) & 
Árið 2010: Suður-Kórea (64,08) > Panama (63,21) - já, Suður-Kórea hæst \\
\addlinespace
In 2015: Suður-Kórea (63.69) > Panama (62.11) & 
Árið 2015: Suður-Kórea (63,69) > Panama (62,11) - já, Suður-Kórea hæst \\
\addlinespace
South Korea has the highest values for all three years. & 
Suður-Kórea er hæst í öllum árum. \\ 
\bottomrule
\end{tabularx}
\caption{Comparison of Claude reasoning for the same question \#11 in Icelandic language across independent runs (same prompt, identical settings).}
\label{tab:claude-comparison-11}
\end{table*}

\begin{table*}
\centering
\footnotesize
\begin{tabularx}{\textwidth}{X X}
\toprule
\textbf{Response A} & \textbf{Response B} \\ 
\midrule
The text clearly states: "Тоғыз ай бойы зерттеген моайды, яғни аралдың мүсіндерін және құздарын бүгін кешке аралап, олармен қоштасамын." & 
Мәтінде профессор өзінің жұмысы туралы қандай ақпарат берген бөлімді іздейміз. \\
\addlinespace
This translates to: "For nine months I have been studying moai, that is, the statues and caves of the island, and today in the evening I will walk around and say goodbye to them." & 
Негізгі сөйлем: "Тоғыз ай бойы зерттеген моайды, яғни аралдың мүсіндерін және құздарын бүгін кешке аралап, олармен қоштасамын." \\
\addlinespace
The blog was published on "23 мамыр" (May 23rd), and the professor states that he has been working there for "тоғыз ай" (nine months) and is now finishing his work. & 
Және сонымен қатар: "Аймақтағы жұмысымды аяқтадым, енді үйге қайтуым керек." \\
\addlinespace
The question asks when the professor started his work in the region according to the blog data. & 
Профессор мәтінінде анық айтады: \par
- Бүгін (23 мамыр) солай жарияланды \par
- Тоғыз ай бойы зерттегенін \par
- Бүгін аралдан кетеленің соңғы күні болғанын \\
\addlinespace
If the professor has been working for nine months and is finishing now (on May 23rd), then he would have started approximately nine months before May 23rd. & 
Осы ақпаратқа сүйене отырып, егер өзі "тоғыз ай бойы" зерттеген болса, ол тоғыз ай бұрын басталған деген сөз. \\
\addlinespace
The answer is clearly stated in the text: the professor worked there for nine months ("Тоғыз ай бойы"). & 
Басқа опцияларды тексерейік: \par
- A: 1990 жылдар - бұл қимасы қаралған мүсіндерді қалай жылжыттығы ашылған уақыт, профессордың жұмысының уақыты емес \par
- C: Бір жыл бұрын - тоғыз ай нақты айтылған \par
- D: Мамыр айының басында - мәтін 23 мамырда жарияланды бірақ жұмыс тоғыз ай бұрын басталды \\
\addlinespace
Therefore, the answer is B. Тоғыз ай бұрын (Nine months ago). & 
Түзік жауап - B: Тоғыз ай бұрын \\ 
\bottomrule
\end{tabularx}
\caption{Comparison of Claude reasoning for the same question \#14 in Kazakh language across independent runs (same prompt, identical settings).}
\label{tab:claude-comparison-14}
\end{table*}

\end{document}